# Improving Mechanical Ventilator Clinical Decision Support Systems with A Machine Learning Classifier for Determining Ventilator Mode

Gregory B. Rehm[a], Brooks T. Kuhn[b], Jimmy Nguyen[b], Nicholas R. Anderson[b], Chen-Nee Chuah[a], Jason Y. Adams[b]

[a]University of California Davis, Davis CA 95616, USA
[b]University of California Davis Medical Center, Sacramento CA 95817, USA

**Abstract**

*Clinical decision support systems (CDSS) will play an increasing role in improving the quality of medical care for critically ill patients. However, due to limitations in current informatics infrastructure, CDSS do not always have complete information on state of supporting physiologic monitoring devices, which can limit the input data available to CDSS. This is especially true in the use case of mechanical ventilation (MV), where current CDSS have no knowledge of critical ventilation settings, such as ventilation mode. To enable MV CDSS to make accurate recommendations related to ventilator mode, we developed a highly performant machine learning model that is able to perform per-breath classification of 5 of the most widely used ventilation modes in the USA with an average F1-score of 97.52%. We also show how our approach makes methodologic improvements over previous work and that it is highly robust to missing data caused by software/sensor error.*

**Keywords:**

Respiration, Artificial; Decision Support Systems, Clinical; Machine Learning

## Introduction

Mechanical ventilation (MV) is a life-saving intervention delivered in the intensive care unit (ICU) to patients with acute respiratory failure. When delivered properly, MV can allow injured lungs heal while the ventilator performs the majority of work of breathing for a patient. When delivered improperly, MV has been associated with a variety of adverse clinical outcomes including patient discomfort, increased sedative dosing, longer ICU length of stay, increased chance of ventilator-induced lung injury, and lower survival [1], [2]. A new generation of clinical decision support systems (CDSS) promises to reduce chances of delivering improper MV by automating aspects of ventilator configuration, and by providing clinically accurate and relevant alerts to providers. However, a key detriment to these systems is that they often lack access to the configured state of the ventilator and therefore lack information that may improve the efficiency of these CDSS [3].

One such piece of information that many MV CDSS lack is the choice of ventilation mode (VM) that determines the pattern of flow and pressure delivery with each breath (Figure 1 B-D). This information is generally unavailable to CDSS due to the lack of interoperability and information exchange between CDSS and the ventilator or electronic health record [3]. CDSS knowledge of VM is important because changing VMs may be a necessary procedure in the course of care for a patient [4]. For example, if CDSS determines that a patient is breathing asynchronously with the ventilator, it may be able to make recommendation that providers choose a different VM that provides more comfort and flexibility in breathing to patients [5]-[8]. Another example would be that CDSS could provide alerts to clinicians if patients continually violate safe volumes of air to inhale. This would be especially important in cases where patients have acute respiratory distress syndrome and need limited tidal volumes [9], [10]. In this case the CDSS could recommend that patients be placed on a VM that limits tidal volumes such as volume-control.

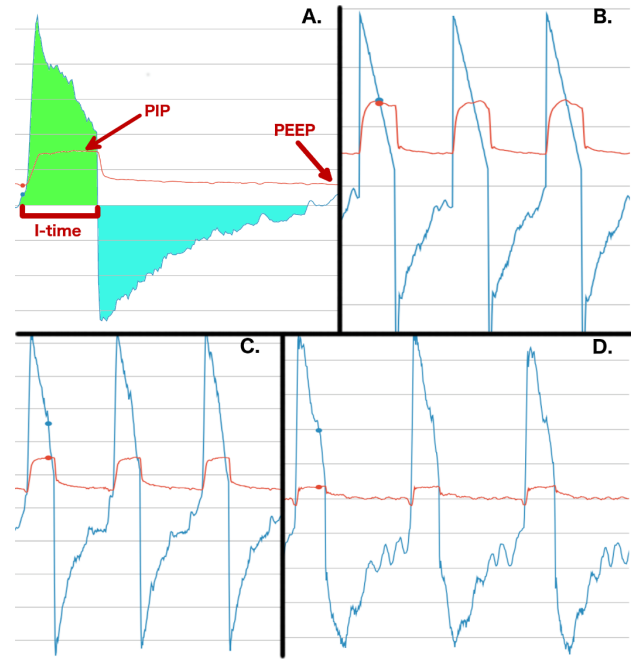

*Figure 1- This figure displays visualizations of ventilator waveform data (VWD). Flow measurements are represented in blue, and pressure in red. A.) Here we display the examples of how to extract information from VWD. Positive End Expiratory Pressure (PEEP) is noted as the minimum pressure supplied by a ventilator, and peak inspiratory pressure (PIP) is the maximum pressure supplied during inhalation. Inspiratory time (I-time) is the amount of time a patient breathes in. Total amount of air breathed in is represented in green, and air breathed out is shown in teal. B.) Shows a canonical example of volume control (VC), a mode where a patient receives a fixed volume of air for each breath. C.) Shows an example of pressure control (PC). In PC pressure is fixed during inhalation. D.) An example of continuous positive airway pressure (CPAP). Here minimal pressure support is given, and all breaths are initiated by the patient.*

If MV CDSS lack knowledge of VM from more traditional methods, it may still be able to access it by utilizing information derived from streams of flow and pressure readings that comprise ventilator waveform data (VWD). To the best of our knowledge, only one previous effort has developed a rule-based classifier using analysis of VWD for providing hourly VM classifications. However, its use of a closed dataset, its limited temporal resolution, and the accuracy of the model represent potential limitations both for research and decision support [3], [5]. Having highly granular temporal resolution VM classification results is important because in practice providers may change VM frequently based on changes in clinical state or patient tolerance of VM. These changes may cause specific VM to remain constant for as low as minutes of time. To improve upon previous work, we note that machine learning (ML) has proven capable of accounting for the highly variable nature of physiologic data such as VWD on temporally granular time scales [11], [12]. So we created a ML model that could identify different VMs on a per-breath basis, with a freely accessible dataset, using only VWD as input.

In this paper, we detail multiple important considerations for modeling an ML classifier that can classify ventilator mode. First we discuss how we created one of the largest datasets of per-breath labeled information, the extraction of features from VWD, and the performance of our resulting ML model that can determine 5 of the most widely used ventilation modes in the USA [4]. Second, we discuss experiments of how well our model performs in the presence of missing training data. Finally, we discuss experimentation we conducted for reducing the size of our training dataset by nearly 72% while maintaining generalizability of our classifier to our testing set. To allow reproducibility of our work, our code and dataset are publicly accessible and published on GitHub[1]. Thus, we hope that our work will serve as a catalyst for continuing to improve the capabilities and efficiency of MV CDSS.

## Methods

In this study, we used a dataset of VWD collected from 103 subjects (IRB# 647002) within intensive care environments of the University of California Davis Medical Center (UCDMC) consisting of MV flow and pressure measurements sampled at 50 Hz [13], [14]. VM was not recorded in the course of VWD data collection. We then randomly selected 2-4 hour epochs of VWD from the 103 subjects. All VWD was stored in data files of 2 hours in length, and approximately 2000 breaths were stored per data file. Each breath in these epochs was then annotated by three expert clinicians (JYA, BTK, JN) for the presence of one of five VMs: volume control (VC), pressure control (PC), pressure support (PS), continuous positive airway pressure (CPAP), and proportional assist ventilation (PAV) (Table 1). Many patients had 2-4 hour periods selected where ventilator mode was switched multiple times, Other modes such as pressure regulated volume control (PRVC), volume support, and airway pressure release ventilation (APRV) were found, and annotated within these epochs, but were excluded in our final analysis because of their rarity of use at UCDMC.

*Table 1- Descriptive statistics for our dataset for each ventilator mode analyzed. We also analyzed number of patient ventilator asynchrony (PVA), suction, and cough breaths found [14]. While these breaths do not represent normal breathing, they are typical in clinical practice.*

|  | Volume Control | Pressure Control | Pressure Support | CPAP | PAV |
|---|---|---|---|---|---|
| Patients | 23 | 37 | 55 | 28 | 22 |
| Total Breaths | 61,662 | 78,635 | 92,360 | 14,795 | 36,303 |
| PVA Breaths | 7,714 | 4,570 | 6,924 | 2,373 | 7,669 |
| Suction Breaths | 750 | 136 | 681 | 350 | 373 |
| Cough Breaths | 229 | 117 | 178 | 56 | 96 |

Because VWD is so heterogeneous it can be difficult for even expert clinicians to make consistent classification of breathing patterns [15]. Thus, in performing classification of VM we ensured that each breath was dual clinician adjudicated, meaning that two clinicians would independently annotate a single breath, and if the classifications disagreed they would be resolved through communication between the two [14]. To further account for breathing heterogeneity, we included regions containing pathologic patient-ventilator interactions such as patient-ventilator asynchrony (PVA), routine clinical events such as suctioning and cough, and regions of noisy data caused by moisture/blood/mucus in the ventilation circuit tubing [14].

*Table 2- The set of proposed features for our model. Features were segmented into per-breath and multi-breath time frames.*

| Feature | Description |
|---|---|
| Inspiratory Flow Slope Variance (per breath) | This feature measures the variance of successive, 0.08-second long slope measurements of the inspiratory flow curve of a single breath. This feature was effective for classifying volume control. |
| Variance of Pressure (per breath) | This feature takes the variance of all pressure measurements for a single breath. This feature was helpful for classifying CPAP which typically utilizes low pressures relative to PEEP on inspiration. |
| Variance of Per-Breath Inspiratory Flow Slope Variance | The inspiratory flow slope variance was found on a per breath basis, and this feature takes the variance of the inspiratory flow slope variance across a 10 breath window. |
| Inspiratory Time (I-time) Variance (10 breath window) | The amount of time that a patient inhales for a single breath is called the I-time. This feature calculated the variance of 10 successive breaths. |
| Pressure-Based I-time Variance (10 breath window) | We defined pressure-based I-time as the amount of time (seconds) that pressure is elevated by [0.4 * (PIP - PEEP)] above PEEP. This was an important variable to measure in pressure control and pressure support, where flow-based I-time can be shorter than the ventilator's set I-time, which may occur in delayed cycling asynchrony. |
| N Plateau Pressures (20 breath window) | A plateau pressure is taken on a ventilator when inspiratory flow is set to 0 for a certain amount of time, during which a ventilator can read the residual pressure in the respiratory system. PAV will repetitively take plateau pressures in order to adjust ventilation to a patient's needs. |
| Pressure-Based I-time Variance (100 breath window) | In this feature, the pressure-based I-time statistic is also calculated for a 100-breath window. This feature was necessary to provide capacity for differentiating between pressure control and pressure support in synchronously breathing patients. |

With this dataset, we utilized 55 patients and 140,928 breaths for our training cohort, and 48 patients and 165,988 breaths for our testing cohort. There was no patient overlap between testing and training cohorts. The testing set was chosen to be approximately as large as the training set because initial modeling yielded strong results, and we wished to utilize a large testing set as further validation for our approach. Using both Scikit-learn and Pytorch ML libraries [16], [17], we then evaluated the use of multiple ML algorithms including: support vector machine (SVM) [18], multi-layer perceptron (MLP), long-short term memory recurrent neural network (LSTM RNN) [19], logistic regression, and a random forest (RF) classifier [20]. All models performed classification on a per-breath basis, the highest possible level of granularity possible in VM classification. Based on model investigation, we settled on usage of the RF with a parameterization of 30 classifier trees for our final model (see online supplemental[2]).

---

[1] https://github.com/hahnicity/ventmode

Our feature set is composed of 7 items of expert-guided information derived from raw VWD, and is described in Table 2. Our features are derived from both per breath and multi-breath analytic time frames. Per-breath time frames occur over a single breath, while multi-breath time frames are composed of windows of short, medium, and long periods of breathing. The short window is 10 breaths long, the medium window is 20 breaths, and the long window is 100 breaths. Tuning of features and hyperparameters was guided by performing 10-fold cross-validation of our training data. After tuning model hyperparameters during the validation phase, we evaluated our model on our testing set. No additional changes to our feature set, or model hyperparameters were performed after model development was completed in the training set. Model performance is primarily reported through F1-score because it is more representative of class-imbalanced classifier performance than accuracy is. F1-score is calculated as the harmonic mean of precision (PPV) and recall (sensitivity):

$$\text{F1-score} = 2 \frac{precision * recall}{precision + recall}$$

A limitation to using RF to classify ventilator mode is the RF classifier assumes that all breaths are independent of each other. However, ventilator mode is a continuous setting that does not vary over time, unless it is manually changed by the provider. Therefore, one breath's mode is often predictive of the next breath's mode. This modeling incongruity causes the RF classifier to sometimes perform an incorrect VM classification even in periods where the classifier correctly predicts the correct VM for a majority of breaths. To smooth these incorrect predictions, we implement an algorithm we term "look-ahead smoothing" which operates as a second pass heuristic on all per breath RF breath predictions. More specifically, once the RF is finished, look-ahead smoothing examines each breath VM classification sequentially, and if it determines a breath's classification is not in accordance with the previous $n$ breaths then it will look ahead at the next $n$ breaths in sequence. The breath will then be re-classified in accordance to the majority $x$ percent of the subsequent $n$ breaths. Both $n$ and $x$ are configurable parameters that we set at $n = 50$ and $x = 60$, parameters which were found via sensitivity analysis. In real-time classification, assuming an average respiratory rate of 20 breaths per minute, this technique results in a latency of at most 2.5 minutes between a given breath and the availability of its final classification.

Finally, we implemented an experiment to test how well our classifier would generalize to a larger dataset if random breaths in our training dataset were missing due to some technical error. So we conduct an experiment where we ablate (i.e. remove) data observations at random from our training dataset in equal proportion for VC, PC, PS, CPAP, and PAV. We do not perform any ablation on the testing set. We then report results of this experiment by recording F1-score for each class with respect to the percentage of the dataset that simulated as missing.

## Results

Using a RF model with the feature set defined in Table 2, we initially performed 10-fold cross validation with our training set to test performance of our VM classifier. We found that during cross validation our model consistently performed within 98-99% for F1-score, recall, and specificity for all VMs. We then evaluated our model on the withheld test set. CPAP suffered the largest drop in performance because it confused PS for CPAP for an entire patient. VC/PAV suffered no drop in performance and PC/PS only suffered slight declines in performance (Table 3).

*Table 3 - Performance of our Random Forest model when applied to our withheld testing set.*

| Mode | F1-Score | Accuracy | Precision | Recall | Specificity |
|------|----------|----------|-----------|--------|-------------|
| VC   | 0.999    | 1.0      | 0.998     | 1.0    | 1.0         |
| PC   | 0.989    | 0.993    | 0.983     | 0.996  | 0.992       |
| PS   | 0.975    | 0.981    | 0.993     | 0.958  | 0.996       |
| CPAP | 0.85     | 0.988    | 0.767     | 0.952  | 0.989       |
| PAV  | 0.994    | 0.999    | 0.99      | 0.998  | 0.999       |

We hypothesized that since the model performed well on both training and testing sets that it would also be robust to scenarios in which breath data went missing due to reason of sensor or software failure. We report results for this experiment in Figure 2. We found the model is robust to missing data until approximately 90% of data is removed. After this point PC and PS F1-score performance begins to decrease and other classifications begin to fluctuate. After 99% of data is removed our classifications lose clinical utility.

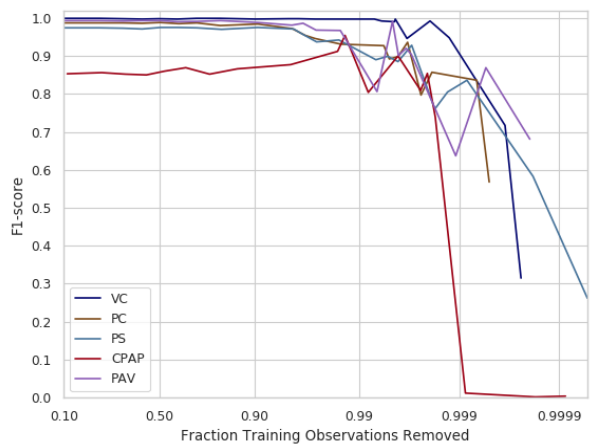

Figure 2- Here we simulate the scenario where a percentage of training observations is missing due to some kind of software/hardware error.

Given the results of the random ablation experiment we hypothesized that we may have created too large a training set. To reduce the size of our training set in a generalizable, non-random way, we hypothesized we only needed to keep the first of a certain number of contiguous breaths from each VM per data file, and still maintain the performance of our original model. In this respect, we could make recommendations to physicians to only annotate the first $m$ breaths in a series and just leave the rest alone. This could also decrease the amount of time necessary to annotate VM on future patients. So, we performed a sensitivity analysis to determine what is the optimal number of contiguous observations to keep per ventilator mode. We do this by sequentially iterating over each VM in our training set and only picking the first $m$ breaths in a file while keeping the number of observations from other VMs constant. Our analysis (Figure 3) showed that it was most optimal to only use the first 450 VC observations, the first 120 PC, 1,200 PS, 160 CPAP, and 80 PAV observations in a file. Using this methodology, we ablated the overall number of training observations by 71.41% from 140,928 to 40,285 observations, while still maintaining generalizability of our training set to our withheld test set, and largely improved CPAP performance (Table 4). By performing ablation we were able to boost the average F1-score of our classifier to 0.9752 from 0.9614 that was reported in Table 3.

---
[2] https://github.com/hahnicity/ventmode/supplemental

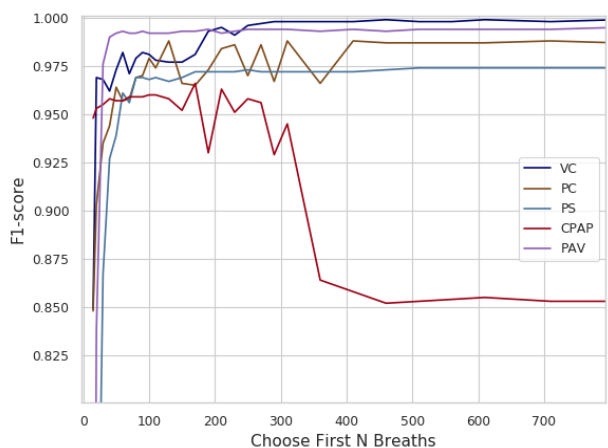

*Figure 3- Results from our sensitivity analysis for choosing the first N contiguous breaths for a given mode in a data file.*

*Table 4- Final results of our ablation experiment where we only keep the first 450 VC, 120 PC, 1,200 PS, 160 CPAP, and 80 PAV observations in a data file. We note the final number of training observations that we kept, and report how much of a reduction that was in contrast to the original training set. Performance improvements/degradation over results listed in Table 3 are bracketed alongside final performance metrics. Eg. a performance increase of 2.0% is denoted as (+.02).*

| Mode | Training Observations | F1-Score |
|------|----------------------|----------|
| VC   | 6,079 (-83.65%)      | 0.998 (-0.001) |
| PC   | 2,154 (-92.77%)      | 0.964 (-0.025) |
| PS   | 27,892 (-26.81%)     | 0.967 (-0.008) |
| CPAP | 3,040 (-73.55%)      | 0.955 (+0.105) |
| PAV  | 1,120 (-94.46%)      | 0.993 (-0.001) |

## Discussion

In this paper, we highlighted how we created a dataset of 308,957 breaths annotated for VM on a per-breath basis and how we developed a highly accurate, ML-based VM classification model that only utilizes raw VWD to perform classifications. Our VM classifier was highly performant in detecting five of the most widely used VMs in the USA, even in the presence of signal noise, episodes of PVA, and routine clinical events such as cough and suction [4]. Using our approach, we were able to achieve methodological and performance improvements in VM classification compared to the current state of the art [3]. In this regard, Murias reported 89% accuracy at detecting per-hour VM classification, and we report average accuracy of 98.05% of per-breath VM classification (Table 3). Finally, we examined how robust our model is to the presence of missing training data, and additional experimental results that suggested how we can decrease the size of our dataset while still maintaining generalizability of our classifier.

We took multiple measures to ensure we were not overtraining our classifier. First, we utilized a relatively large and diverse sampling of patients to create both our testing and training sets. This created one of the largest available datasets of per-breath labeled VWD. 2-4 hour epochs were chosen at random from each of these patients. Our testing set included almost as many patients as our training set, and was composed of more breaths than our training set. There was no overlap of patients between training and testing sets. Finally, all model features and hyperparameters were evaluated on the training set using K-fold validation, and were frozen after initial evaluation of our testing set.

Our ablation experiments deserve additional consideration. The results of the random ablation experiments highlight multiple things: 1) RF is extremely performant with our featurization approach, and is also able to perform VM classification with small amounts of data. 2) Our ablation results also illustrate that it may not be necessary to create very large training datasets of information to create performant ML classifiers for VM. 3) Our size reduction experiments did see some decreases in performance in PC and PS because of the manner in which we performed our sensitivity analysis. In our analysis we only modified observations from a single VM type while keeping other VM observations constant, so it was not possible to determine side effects from simultaneously ablating several modes at once. Future experiments could perform the more computationally demanding task of ablating multiple modes at once to further explore the issue. 4) Our size reduction experiment showed that the first 160 breaths seem to be most representative of CPAP breathing patterns. We hypothesize this can be explained by the fact that some patients tire quickly when on CPAP, and thus their breathing can become more irregular. In this case, later breaths in CPAP sequences may more closely resemble asynchronous breathing from other ventilator modes instead of CPAP.

This work had several limitations. Our use of "look-ahead smoothing" introduced a small latency of 2.5 minutes to real-time ventilator mode predictions. This time delay in classification would not likely be of clinical consequence since CDSS recommendations requiring VM state information would rarely be executed over such short time frames to ensure that transient changes in waveforms do not trigger frequent false alarms. If latency is not desired then "look-behind smoothing" can be used as an alternative approach. Our study was also confined to a single academic medical center and single ventilator type. There are also additional ventilator modes such as PRVC that we were unable to add to our model due to their paucity of use at UCDMC. We welcome additional collaboration and inclusion of multi-center data and have publicly released our code and dataset.

## Conclusions

In conclusion, we created a highly-performant ML classifier for detecting five of the most commonly used ventilator modes in the USA, using only raw VWD as input. Our use case further demonstrates the utility of ML analysis of physiologic waveform data to improve CDSS characterization of patient state when state is missing due to limitations of available informatics infrastructure. We also illustrated the usage of dataset ablation to characterize how missing data affects the generalization performance of our classifier, and how we can restrict size of our training set while maintaining model generalization to our test dataset. Our classifier will facilitate development of more advanced automated MV CDSS to improve the management of patients experiencing respiratory failure.

## Acknowledgements

This work was generously supported by the National Heart, Lung, and Blood Institute (NHLBI) Individual Predoctoral Fellowship (award number 1F31HL144028-01), and NHLBI Emergency Medicine K12 Clinical Research Training Program (grant number K12 HL108964)

**Address for correspondence**

For correspondence: email Gregory B. Rehm: grehm@ucdavis.edu